**Title:**

**Assessing the Quality of AI-Generated Clinical Notes: A Validated Evaluation of a Large Language Model Scribe**

**Authors:**

1)      Erin Palm, MD MBA FACS

[1] Suki AI, Redwood City, CA

[2] Santa Clara Valley Medical Center, San Jose, CA

[3] Department of Surgery, Division of General Surgery, Stanford University School of Medicine, Stanford, CA

2)      Astrit Manikantan, MBA

Suki AI, Redwood City, CA

3)      Mark E. Pepin, MD, MS, PhD

[1] Stanford Cardiovascular Institute, Stanford University School of Medicine, Stanford, CA

[2] Department of Medicine, Division of Cardiovascular Medicine, Stanford University School of Medicine, Stanford, CA

4)      Herprit Mahal, MD FACP

[1] Suki AI, Redwood City, CA

[2] Hippocratic AI, Palo Alto, CA

[3] The Permanente Medical Group, Oakland, CA

5)      Srikanth Subramanya Belwadi, MBA

Suki AI, Redwood City, CA

**Abstract**

In medical practices across the United States, physicians have begun implementing generative artificial intelligence (AI) tools to perform the function of scribes in order to reduce the burden of documenting clinical encounters. Despite their widespread use, no established methods exist to gauge the quality of AI scribes. To address this gap, we developed a blinded study comparing the relative performance of large language model (LLM) generated clinical notes with those from field experts based on audio-recorded clinical encounters. Quantitative metrics from the Physician Documentation Quality Instrument (PDQI-9) provided a framework to measure note quality, which we adapted to assess relative performance of AI-generated notes. Board-certified experts spanning 5 medical specialties used the PDQI-9 tool to evaluate specialist-drafted "Gold" notes and LLM-authored "Ambient" notes. Two evaluators from each specialty scored notes drafted from a total of 97 patient visits. We found uniformly high inter-rater agreement (RWG > 0.7) between evaluators in general medicine, orthopedics, and obstetrics / gynecology, and moderate (RWG 0.5-0.7) to high inter-rater agreement in pediatrics and cardiology. We found a modest - yet significant - difference in the overall note quality, wherein Gold notes achieved a score of 4.25/5 and Ambient notes scored 4.20/5 ($p = 0.04$). Our findings support the use of the PDQI-9 instrument as a practical method to gauge the quality of LLM authored notes, as compared to human-authored notes.

**Background**

Physicians and health systems have begun rapidly adopting software applications that employ large language models (LLMs) to facilitate note writing for patient encounters.[1–3] Available software performs functions similar to those of medical scribes, which has been shown to increase physician satisfaction and productivity,[4] but at lower cost and with greater scalability. In principle, any scribe software is most beneficial when it generates a draft note of high quality, because the physician must review and edit the draft before finalizing the note for the medical record. Despite its numerous advantages, the introduction of LLM-generated clinical notes raises questions about documentation quality particularly given the field-specific needs and expectations of the medical record. Prior literature has established at least two validated instruments for evaluating physician note quality, including PDQI-9[5] and Q-Note[6]. However, these tools have not yet been systematically applied to LLM-generated notes and assessed via specialist review.

Ambient scribes refer to LLMs that passively collect and interpret conversations to extract meaningful and structured content, allowing the clinician to focus on the patient interaction. Our group has worked to optimize such an Ambient scribe, Suki, which has been designed to summarize clinical encounters into structured notes using audio-recorded medical interactions. This scribing function involves 3 major steps: integrating with the physician's electronic health record (EHR) to retrieve information about the patient and encounter context, transcribing the conversation between the doctor, patient, and any other visit participants, and then creating a summary of this information, as appropriate for the Suki user's specialty, in the form of a clinical note. Both proprietary and tuned third-party language models are used to perform this functionality.

In the current study, we compare the relative performance of Suki's Ambient scribe-generated clinical notes with those drafted by board-certified specialists spanning 5 clinical domains, which were then reviewed by blinded experts in the respective fields.

**Methods**

We retrospectively queried encounters from Suki's production database from October 2024 to identify encounters representing the following clinical specialties: general medicine, pediatrics, obstetrics and gynecology, orthopedic surgery, and cardiology. Audio recordings were retrieved for the selected encounters and transcribed using automated speech recognition (ASR) software. To ensure that all audio recordings were de-identified, a team of operations specialists systematically reviewed both the audio files and their transcripts, selecting only encounters that contained no personal health information (PHI) that could potentially identify patients. We also excluded encounters with audio duration less than 1 minute, very poor audio quality which impaired the ability to produce a transcript, and visits conducted in a non-English language. Out of 930 visits screened, 126 were selected as eligible. Of these, we assigned the first 20 qualifying encounters for each specialty, reserving the remaining visits as back-ups. The software then provided the visit transcript, along with limited information about the patient and clinician, to an LLM to create an "Ambient" note for each encounter. In OB/Gyn, 3 visits were later excluded due to the patient audio not being recorded during telehealth visits.

*Note reviewers.* To provide reference documentation for each encounter, physician specialists were recruited from the respective medical field to draft notes, termed "Gold notes." The Gold note author had access to the same inputs as the LLM, including the audio recording, transcript, and associated patient and clinician information. No Gold note authors were directly involved in the index patient encounter to ensure that both the Gold and Ambient notes were generated from identical source material.

*Quality assessments.* Two board-certified clinicians were recruited to perform evaluations of the Ambient and Gold notes. As with the Gold note physician authors, the evaluators represented the appropriate medical specialties and included a mix of board-certified physicians, fellows, residents, and physician assistants. Evaluators had access to the encounter audio, transcript, and information about patient and clinician, and were blinded to the origin of the two notes – they were asked to evaluate notes written by "Model 1" and "Model 2."

The evaluators rated each note according to the criteria described in Table 1. These criteria were largely drawn from the PDQI-9 instrument, which we chose over Q-Note because PDQI-9 can be applied more flexibly in a variety of clinical settings. The PDQI-9, importantly, was "not designed to assess the presence or absence of specific note components (e.g. "reason for admission" in an admission note)" to ensure broad applicability of the instrument. Instead, it uses subjective Likert scale ratings by physicians, similar to other frameworks proposed for the evaluation of LLMs in healthcare.[7] A validation exercise for the PDQI-9 was previously performed using internal medicine admission notes (Stetson et al. 2012).

***Quality Assessment.*** For our note quality assessment, we adopted 8 of the 9 questions from the original PDQI-9. We removed the "Up to date" criterion due to its not being applicable to the AI scribe use case. We added "Appropriate for specialty" and "Fair" as additional criteria, as well as a qualifier for the "Accuracy" criterion that identifies whether any inaccuracies were due to "Hallucination." Each criterion was rated on Likert scale from 1 to 5, where 5 = Extremely and 1 = Not at all, except the Hallucination criterion which was either "Present" or "Absent." The evaluator also provided their Overall Preference between the 2 notes by specifying "Model 1," "Model 2," or "I prefer both equally."

**Table 1**: Evaluation criteria for clinical note quality.

| Criterion | Meaning |
|---|---|
| Accurate | The note is true. It is free of incorrect information. |
| Hallucination | Are any inaccuracies due to hallucinated content? (binary) |
| Thorough | The note is complete and documents all the issues of importance to the patient. |
| Useful | The note is extremely relevant, providing valuable information and/or analysis. |
| Organized | The note is well-formed and structured in a way that helps the reader understand the patient's clinical course. |
| Comprehensible | The note is clear, without ambiguity or sections that are difficult to understand. |
| Succinct | The note is brief, to the point, and without redundancy. |
| Synthesized | The note reflects the author's understanding of the patient's status and ability to develop a plan of care. |
| Internally Consistent | No part of the note ignores or contradicts any other part. |
| Appropriate for Specialty | The language and content of the note is typical for this medical specialty. |
| Fair | The note does not display prejudice based on race, gender, or other aspects of the patient's identity. |

***Statistical analysis.*** For all pairwise comparisons, the Shapiro-Wilk test for normality was performed to determine the most appropriate statistical test. All pairwise factors exhibiting a parametric distribution were evaluated using the student's t-test with Benjamini-Hochberg adjustment; otherwise, a Mann-Whitney test was used. All data are reported as mean ± standard deviation unless otherwise specified. Statistical analyses and data visualization were completed using GraphPad Prism version 10.4.0 for Macintosh (GraphPad Software, San Diego, CA) and R software, version 4.4.2 (R Foundation for Statistical Computing, Vienna, Austria). Statistical significance was assigned when $P < 0.05$ unless otherwise specified. Within-group agreement was assessed using the RWG score, which quantifies consensus among raters by comparing observed variance to expected random variance, as previously described [8].

**Results**

To better characterize the relative performance of our AI-generated medical dictation platform, termed "Ambient" notes, compared to those written by clinical experts, audio and transcripts from 97 patient visits were included in the current study. Experts from five medical specialties, listed in Table 2, drafted notes based on the 20 patient encounters (17 in OB/Gyn) within their field, termed "Gold" notes. For each visit, a Gold and an Ambient note was scored, each by 2 clinical expert evaluators, yielding a sample of 388 notes for analysis.

**Table 2:** Number of visits and evaluations by specialty.

| Specialty | Visits | Evaluations |
|---|---|---|
| General Medicine | 20 | 40 |
| OB / Gyn | 17 | 34 |
| Ortho | 20 | 40 |
| Peds | 20 | 40 |
| Cardio | 20 | 40 |
| **Total** | **97** | **194** |

Agreement between the 2 evaluators was uniformly high (RWG >0.7) across all criteria in general medicine, OB/Gyn, and orthopedic notes. In pediatrics, there was moderate inter-rater agreement (RWG 0.5-0.7) for 4 of 11 criteria, and other criteria had high agreement. In cardiology, there was moderate agreement for 6 of 11 criteria, and poor inter-rater agreement (RWG <0.5) for the Organized criterion.

**Table 3:** Inter-rater agreement by quality criterion

| Specialty | AVG | Accurate | Thorough | Useful | Organized | Comprehensible | Succinct | Synthe-sized | Internally Consistent | Appropriate for Specialty | Fair |
|---|---|---|---|---|---|---|---|---|---|---|---|
| General Medicine | 0.83 | 0.81 | 0.74 | 0.86 | 0.73 | 0.84 | 0.80 | 0.84 | 0.83 | 0.85 | 0.94 |
| OB / Gyn | 0.86 | 0.85 | 0.88 | 0.84 | 0.74 | 0.84 | 0.81 | 0.85 | 0.92 | 0.76 | 0.99 |
| Peds | 0.76 | 0.63 | 0.72 | 0.60 | 0.68 | 0.83 | 0.82 | 0.58 | 0.74 | 0.82 | 1.00 |
| Cardio | 0.68 | 0.79 | 0.57 | 0.61 | 0.48 | 0.62 | 0.54 | 0.73 | 0.71 | 0.63 | 0.92 |
| Ortho | 0.88 | 0.85 | 0.87 | 0.86 | 0.90 | 0.84 | 0.83 | 0.87 | 0.87 | 0.83 | 0.99 |

Average scores across all notes and between the 2 evaluators within each specialty are listed in Table 4 for each the modified PDQI-9 criteria. There was a statistically significant preference for Gold notes over Ambient notes on the Accurate (p = 0.05), Succinct (p < 0.001), and Internally Consistent (p = 0.004) criteria. Ambient notes were preferred over Gold notes on the Thorough (p < 0.001) and Organized (p = 0.03) criteria. For other criteria (Useful, Comprehensible, Synthesized, Appropriate for Specialty, and Fair) the differences were not statistically significant. An overall average of score across all of the modified PDQI items slightly favored Gold notes at 4.25, versus 4.20 for Ambient notes (p = 0.04).

**Table 4:** Note Quality Scores for Gold Notes versus Ambient Notes

| Criterion<br>5=Extremely, 1=Not at all | Gold Note | Ambient Note | Difference | p-value |
|---|---|---|---|---|
| Accurate | 4.13 | 3.98 | -0.15 | 0.05 |
| Thorough | 3.80 | 4.22 | 0.43 | <0.001 |
| Useful | 4.03 | 4.05 | 0.02 | 0.80 |
| Organized | 4.01 | 4.19 | 0.18 | 0.03 |
| Comprehensible | 4.19 | 4.26 | 0.06 | 0.38 |
| Succinct | 4.40 | 3.72 | -0.67 | <0.001 |
| Synthesized | 4.22 | 4.09 | -0.14 | 0.07 |
| Internally Consistent | 4.47 | 4.31 | -0.16 | 0.004 |
| Appropriate for Specialty | 4.38 | 4.29 | -0.09 | 0.24 |
| Fair | 4.82 | 4.83 | 0.01 | 0.70 |
| **Overall Average** | **4.25** | **4.20** | **-0.05** | **0.04** |

*P-value is for a paired, 2-tailed student t test for n = 194 head-to-head evaluations

With respect to Hallucinations, evaluators identified Hallucinations in both Gold and Ambient notes, with the presence of Hallucination identified in 20% of Gold notes, versus 31% of Ambient notes (p = 0.01). The average RWG score for the binary Hallucination criterion was 0.94, confirming high inter-rater agreement for this question.

Analysis of per-specialty average scores for each quality criterion showed that all specialties agreed directionally that the Ambient notes were more Thorough, although this difference was only statistically significant for Cardiology and Pediatrics (Figure 1). In general, for all other criteria, OB/Gyn and pediatrics specialists tended to favor the Gold notes, whereas general medicine, orthopedics, and cardiology specialists tended to favor the Ambient notes.

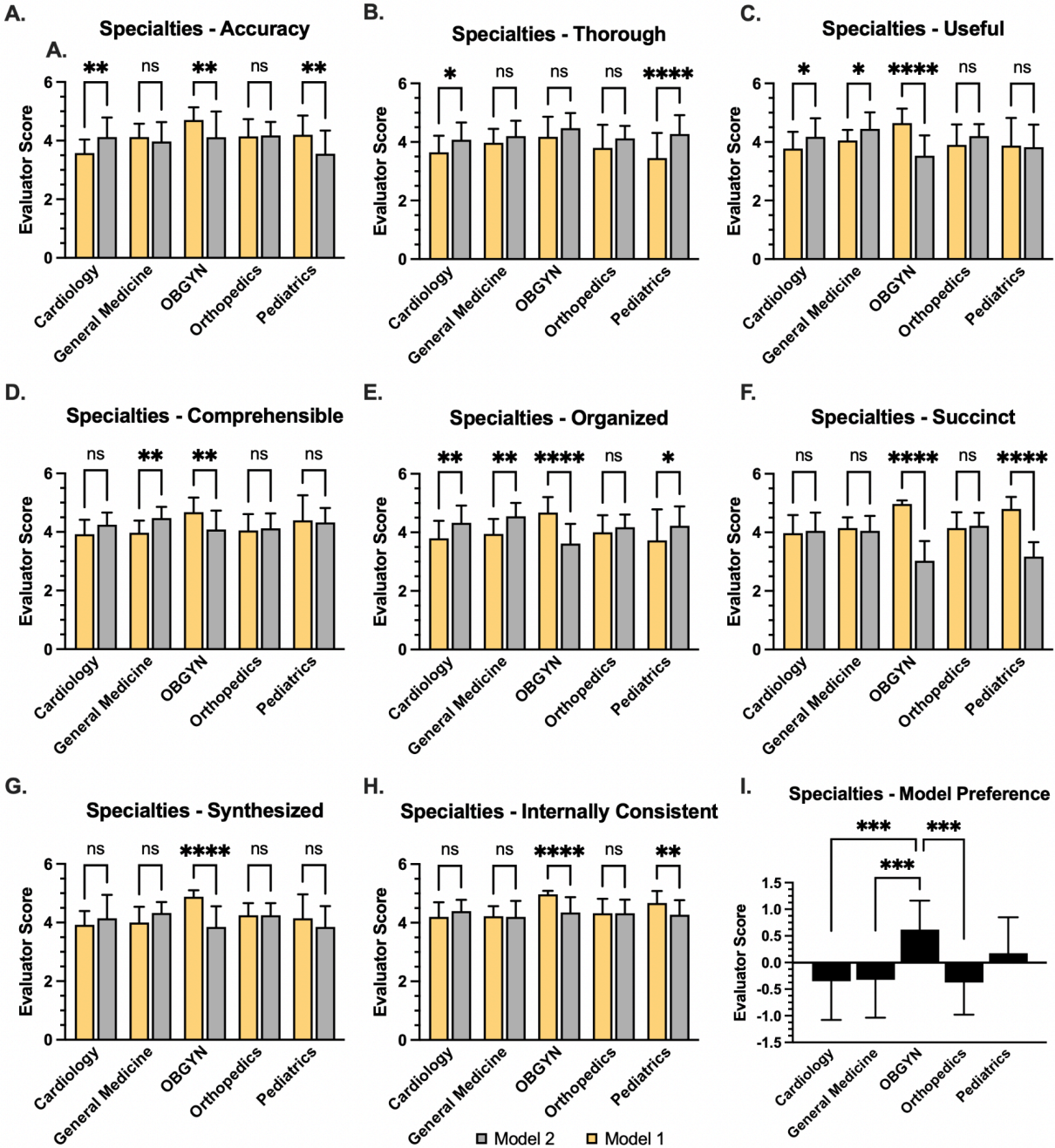

**Figure 1: Relative evaluations for AI-generated notes across medical specialties. (A)** Bar graph of Ai-generated "Ambient" (Model 1) note relative to reference "Gold" note (Model 2) according to Likert scale, also showing **(B)** Thoroughness, **(C)** Usefulness, **(D)** Comprehensibility, **(E)** Organization, **(F)** Succinctness, **(G)** degree of synthesis, and **(H)** internal consistency.

The specialty preference seen in the modified PDQI-9 criteria was consistent with the single-question response from evaluators regarding their overall note preference. Both evaluators in OB/Gyn and orthopedics favored the Gold note more often than the Ambient note, whereas both evaluators in general medicine, orthopedics, and cardiology favored the Ambient note more often.

**Table 5: Overall Note Preference**

Question: Which model note do you prefer? Select: Model 1, Model 2, or I prefer both equally

|  | Gold | Ambient | Both |
|---|---|---|---|
| Overall Preference | 74 | 89 | 26 |
| % | 39% | 47% | 14% |

| Evaluator | Gold Note Preferred | Ambient Note Preferred | Difference |
|---|---|---|---|
| General Medicine 1 | 30% | 45% | 0.15 |
| General Medicine 2 | 10% | 60% | 0.50 |
| OB / Gyn 1 | 88% | 6% | -0.82 |
| OB / Gyn 2 | 65% | 29% | -0.35 |
| Orthopedics 1 | 15% | 50% | 0.35 |
| Orthopedics 2 | 25% | 60% | 0.35 |
| Pediatrics 1 | 50% | 45% | -0.05 |
| Pediatrics 2 | 60% | 35% | -0.25 |
| Cardiology 1 | 15% | 70% | 0.55 |
| Cardiology 2 | 35% | 50% | 0.15 |

Across the sample overall, the responses to the Overall Note Preference question favored Ambient notes more often (Table 5). This preference points in the opposite direction of the scores for all of the PDQI averaged together, which favored the Gold notes (4.25 vs 4.20 for Ambient notes, $p = 0.04$). There are differences between specialty average ratings, as seen in Figure 1, that diverge across the different specialties and are not seen in the overall averages for the combined sample.

**Discussion**

The introduction of automated note-writing and clinical summary tools has the potential to streamline the administrative burden facing clinicians by generating clinical documents that accurately summarize patient encounters. However, the clinical precision of LLM-generated notes is limited by several universal pitfalls and thus requires a standardized validation process. In the current study, for the purposes of quality evaluation and improvement, we adopted an externally validated note quality instrument to compare Gold notes created by specialty experts to Ambient notes generated by an LLM.

By implementing a peer-review process of comparing Ambient and Gold clinical notes, we have established a repeatable method for understanding note quality and identifying areas for improvement. The finding that Gold notes did not consistently perform better than Ambient notes was unexpected. Although on comparison of the global average scores the Gold notes scored slightly better, on some criteria such as Thorough, Ambient notes out-performed the human-authored notes. This implies the Ambient notes did a better job of capturing all the details of the discussion. On other criteria such as Succinct, Gold notes were more highly rated. This reflects the tendency of the LLM author to be more verbose in its writing style. Further qualitative research is required to understand physician preferences with respect to the Thorough and Succinct criteria in particular.

Strengths of this study include adopting a previously validated instrument, the PDQI-9, which applies well to this generative AI use case with few modifications. In analyzing inter-rater agreement for both the externally validated PDQI criteria and the 3 criteria that were added, we found generally high levels of agreement. We did find a lower level of agreement between our cardiology evaluators, and on further exploration we found that one of the cardiology evaluators

Page 13was a "hard grader" – their scores were consistently lower than the other evaluator's scores. In the context of our comparison between Gold and Ambient notes, which used the average of the two evaluators' scores to compare note quality, the "hard grader" issue is not a limitation. However, it limits the ability to compare scores across specialties. In future, mitigations such as evaluator training or score normalization can be explored.

In terms of weaknesses, a primary limitation of this study is the use of a single Gold note author for each specialty. This limits our ability to compare AI scribe performance across specialties, because we are unable to separate the effect of the Gold note author's writing ability from the specialty context itself. In OB/Gyn in particular, the average rating across all criteria for Gold notes was excellent at 4.75 out of 5. We do not know to what extent the excellence of that Gold note author leads to a less favorable comparison with the LLM-generated notes, but the opportunity for quality improvement is clear – whether it's a specialty-specific or note author-specific effect, or both.

Other limitations include the retrospective study design, the possible sample bias introduced by selecting only PHI-free visit audio, and the fact that notes were authored by physicians who were not involved in the patient encounter. With regard to note authorship, the study was designed to create an apples-to-apples comparison of notes written with the same data inputs available to the human author and the LLM author. However, the note by the physician actually involved in the encounter – absent time pressure, if asked to write an ideal note – would likely be considerably higher quality than a note authored using the methodology described here. This should be acknowledged as a limitation of AI scribes in general: the LLM does not have access to all of the rich inputs available to the bedside clinician. Further research is required to better understand these note quality differences given the constrained set of information the LLM has access to.

**Conclusion**

Page 14This study has demonstrated how a previously validated instrument for evaluating note quality, the PDQI-9, can be used with minor adaptations to evaluate quality of LLM-generated clinical notes. It further establishes a methodology for comparing the quality of physician-authored notes to LLM-authored notes via expert clinical review. As expected, physician-authored notes out-performed LLM-authored notes overall, although LLM-authored notes were found to be more Thorough and Organized. More important than the numeric quality results for this static dataset, which includes notes authored in October 2024 and has already become outdated given the pace of change in LLM technology, we have established a methodology which developers can leverage to identify opportunities for quality improvement in LLM generation of clinical notes.

**Data Access**

The dataset on which this manuscript is based, including de-identified transcripts, notes, and annotations, will be available for public access at https://suki.ai/research upon publication of this manuscript. To access the data, the user's name, affiliation, and a valid email address are required, and the user must agree to access the data for the purpose of academic research only.

**References**

1. Shah SJ, Devon-Sand A, Ma SP, et al. Ambient artificial intelligence scribes: physician burnout and perspectives on usability and documentation burden. J Am Med Inform Assoc [Internet] 2025;32(2):375–80. Available from: http://dx.doi.org/10.1093/jamia/ocae295

2. Cain CH, Davis AC, Broder B, et al. Quality assurance during the rapid implementation of an AI-assisted clinical documentation support tool. NEJM AI [Internet] 2025 [cited 2025 Apr 13];2(4). Available from: http://dx.doi.org/10.1056/aics2400977

3. Stults CD, Deng S, Martinez MC, et al. Evaluation of an ambient artificial intelligence documentation platform for clinicians. JAMA Netw Open [Internet] 2025;8(5):e258614. Available from: http://dx.doi.org/10.1001/jamanetworkopen.2025.8614

4. Gidwani R, Nguyen C, Kofoed A, et al. Impact of Scribes on Physician Satisfaction, Patient Satisfaction, and Charting Efficiency: A Randomized Controlled Trial. Ann Fam Med [Internet] 2017;15(5):427–33. Available from: http://dx.doi.org/10.1370/afm.2122

5. Stetson PD, Bakken S, Wrenn JO, Siegler EL. Assessing Electronic Note Quality Using the Physician Documentation Quality Instrument (PDQI-9). Appl Clin Inform [Internet] 2012;3(2):164–74. Available from: http://dx.doi.org/10.4338/aci-2011-11-ra-0070

6. Burke HB, Hoang A, Becher D, et al. QNOTE: an instrument for measuring the quality of EHR clinical notes. J Am Med Inform Assoc [Internet] 2014;21(5):910–6. Available from: http://dx.doi.org/10.1136/amiajnl-2013-002321

7. Tam TYC, Sivarajkumar S, Kapoor S, et al. A framework for human evaluation of large language models in healthcare derived from literature review. NPJ Digit Med [Internet] 2024;7(1):258. Available from: http://dx.doi.org/10.1038/s41746-024-01258-7

8. James LR, Demaree RG, Wolf G. Estimating within-group interrater reliability with and without response bias. J Appl Psychol [Internet] 1984;69(1):85–98. Available from: http://dx.doi.org/10.1037/0021-9010.69.1.85